\documentclass[letterpaper]{article} 
\usepackage{aaai2026}  
\usepackage{times}  
\usepackage{helvet}  
\usepackage{courier}  
\usepackage[hyphens]{url}  
\usepackage{graphicx} 
\usepackage{natbib}  
\usepackage{caption} 
\usepackage{algorithm}
\usepackage{algorithmic}

\usepackage{newfloat}
\usepackage{listings}
\DeclareCaptionStyle{ruled}{labelfont=normalfont,labelsep=colon,strut=off} 
\floatstyle{ruled}
\newfloat{listing}{tb}{lst}{}
\floatname{listing}{Listing}
\title{CoRA: A Collaborative Robust Architecture with Hybrid Fusion \\ for Efficient Perception}
\author{
    Gong Chen\textsuperscript{\rm 1},
    Chaokun Zhang\textsuperscript{\rm 1}\equalcontrib,
    Pengcheng Lv\textsuperscript{\rm 2},
    Xiaohui Xie\textsuperscript{\rm 3}
}
\affiliations{
    \textsuperscript{\rm 1}College of Intelligence and Computing, Tianjin University, Tianjin, China \\
    \textsuperscript{\rm 2}School of Future Technology, Tianjin University, Tianjin, China \\
    \textsuperscript{\rm 3}Department of Computer Science and Technology, Tsinghua University, Beijing, China \\
    \{gongchen01, zhangchaokun\}@tju.edu.cn

}

\usepackage{bibentry}
\usepackage{amsmath} 
\usepackage{amsfonts}
\usepackage{booktabs} 
\usepackage{subcaption}

\usepackage{pifont}
\usepackage{newunicodechar}
\newunicodechar{✓}{\ding{51}}
\newunicodechar{✗}{\ding{55}}

\begin{document}

\maketitle

\begin{abstract}

Collaborative perception has garnered significant attention as a crucial technology to overcome the perceptual limitations of single-agent systems. Many state-of-the-art (SOTA) methods have achieved communication efficiency and high performance via intermediate fusion. However, they share a critical vulnerability: their performance degrades under adverse communication conditions due to the misalignment induced by data transmission, which severely hampers their practical deployment. To bridge this gap, we re-examine different fusion paradigms, and recover that the strengths of intermediate and late fusion are not a trade-off, but a complementary pairing. Based on this key insight, we propose CoRA, a novel collaborative robust architecture with a hybrid approach to decouple performance from robustness with low communication. It is composed of two components: a feature-level fusion branch and an object-level correction branch. Its first branch selects critical features and fuses them efficiently to ensure both performance and scalability. The second branch leverages semantic relevance to correct spatial displacements, guaranteeing resilience against pose errors. Experiments demonstrate the superiority of CoRA. Under extreme scenarios, CoRA improves upon its baseline performance by approximately 19\% in AP@0.7 with more than 5x less communication volume, which makes it a promising solution for robust collaborative perception.
\end{abstract}


\section{Introduction}

Autonomous driving perception systems are fundamentally limited by occlusions and finite sensing ranges, which pose critical safety risks \cite{kekaoxing}. Enabled by Vehicle-to-Everything (V2X) communication, collaborative perception presents a powerful solution by sharing information across agents \cite{Survey1}. However, the benefits of V2X are constrained by real-world communication bottlenecks, such as limited bandwidth and unreliability \cite{siddiqui20215ginterference}. To mitigate these overheads, the research community has converged on intermediate fusion, an approach that exchanges feature-level data. This paradigm has become the prevailing choice for SOTA methods\cite{hu2022where2comm, zhang2024ermvp}.

Existing communication-efficient methods focus mainly on selectively transmitting \cite{hu2022where2comm, yang2023how2comm} or compressing features \cite{wang2023core} to reduce communication overhead. However, a fundamental limitation of these frameworks lies in their presupposition of accurate alignment in a shared coordinate system, which critically depends on precise localization \cite{yazgan2024survey2}.

\begin{figure}[!t]
    \centering
    \includegraphics[width=0.43\textwidth]{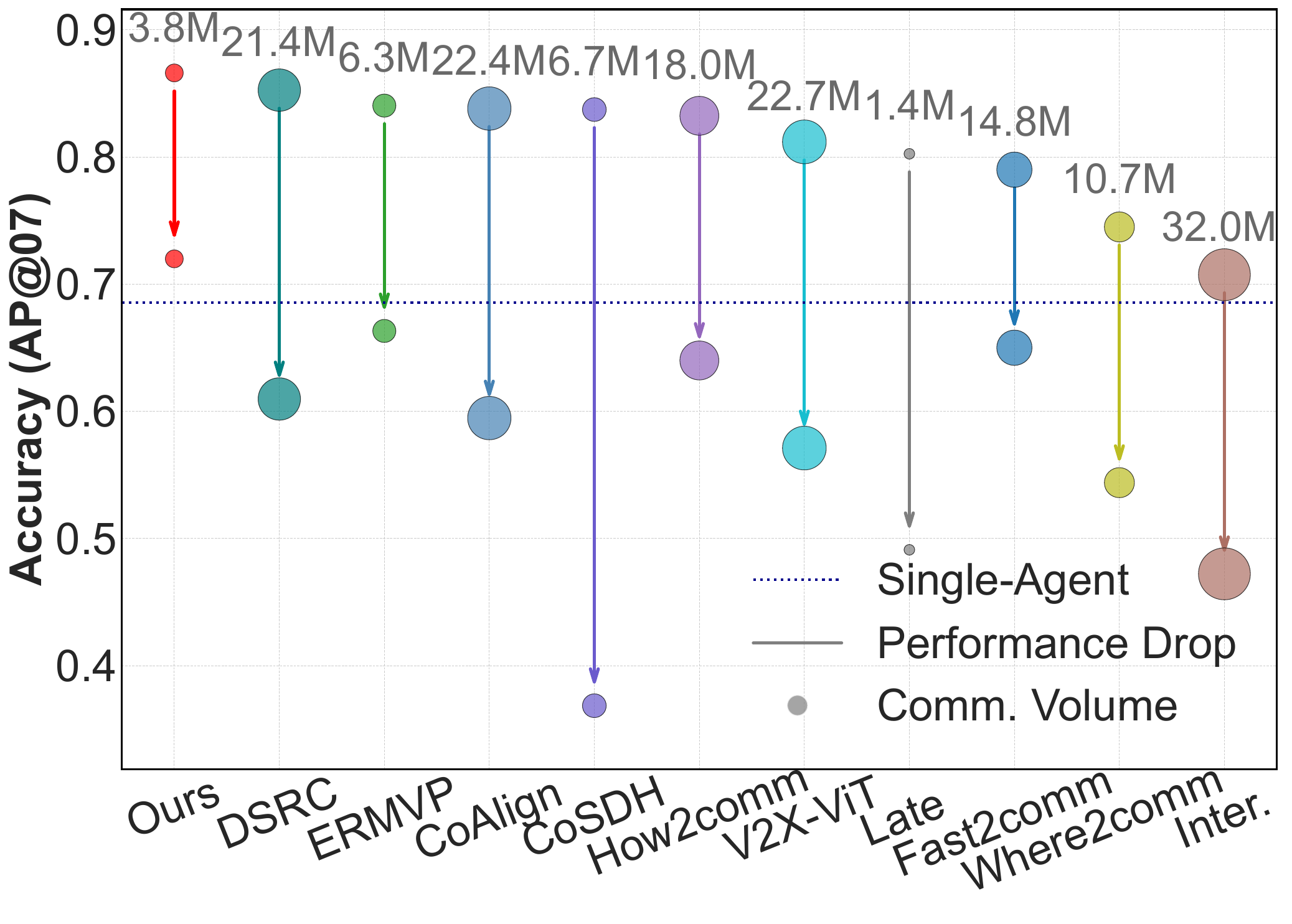}
    \caption{Performance comparison of collaborative perception methods in communication efficiency and robustness.}
    \label{fig:abstract}
\end{figure}

This assumption crumbles when unstable communication channels undermine the integrity of transmitted pose information, which in turn exacerbates pose errors and leads to severe data misalignment during multi-agent fusion \cite{tang2025rocooper}. This highlights a challenge that extends beyond mere bandwidth constraints \cite{su2023uncertainty}. The presence of pose errors makes these efficient strategies susceptible to degradation. The severity of this issue is illustrated in Fig.\ref{fig:abstract}: leading efficient methods degrade in performance under pose errors, falling even below single-agent detection.

To develop a robust solution for these real-world challenges, we first re-examine the vulnerability patterns of different fusion paradigms. Intermediate Fusion leverages dense feature-level interaction to achieve a high performance ceiling when poses are accurate \cite{xu2022v2xvit}. However, any pose-induced misalignment at the feature level can trigger mismatched interactions, resulting in an irrecoverable collapse of the shared representation. In contrast, late fusion exhibits greater resilience. Pose errors do not corrupt the features themselves but affect object association \cite{late}. While this leads to the mislocalization of collaborator detections, the intrinsic integrity of each individual detection is preserved. Although its accuracy ceiling is typically lower, the integrity of information grants it superior rectifiability. Our analysis thus reveals a key insight: \textit{the weaknesses of intermediate and late fusion are complementary.} This discovery provides the theoretical foundation for our hybrid design.

Motivated by this insight, we introduce CoRA, a parallel, dual-branch framework that unifies the complementary benefits of both intermediate and late fusion. This design decouples the pursuit of high performance from pose error robustness, overcoming the apparent trade-off prevalent in conventional methods under strict bandwidth limitations. Specifically, our contributions are embodied in two synergistic branches: 1) A high-performance feature-level fusion branch. We first introduce a receiver-centric Competitive Information Transmission (CIT) module to select and transmit only the most critical features. The module exchanges lightweight confidence maps to identify the optimal information provider for each spatial region and then requests only the critical features to fill its perceptual gaps. This approach keeps the communication overhead nearly constant regardless of the number of collaborators, thereby ensuring the system's scalability. Next, we employ a Lightweight Collaboration (LC) module to enable efficient knowledge sharing across collaborators at the feature level for high-accuracy perception. Its core objective is to exploit the high-performance ceiling of intermediate fusion. The entire process is optimized end-to-end via a training-only guidance module, ensuring that the most valuable features are processed to ultimately maximize the global perception gain.
2) A robust object-level correction branch. A naive combination of fusion strategies is insufficient \cite{xu2025cosdh}, as standard late fusion itself degrades severely under pose errors. To guarantee resilience against pose errors, we introduce a novel Pose-Aware Correction (PAC) module. This module mitigates spatial misalignment by leveraging the semantic relevance between ego and collaborator detections, thereby ensuring stable performance even under pose errors.

Extensive experiments on two benchmarks validate the superiority of our method. On OPV2V, for instance, results show that CoRA achieves a remarkable 86.58\% in AP@0.7 with only 3.80 MB communication overhead under extremely adverse conditions. Compared to the baseline, CoRA increases the detection accuracy by 19.02\% while slashing communication volume by 82.2\%, demonstrating an exceptional balance between efficiency and robustness. The contributions of this work are summarized as follows:
\begin{itemize}
\item We propose CoRA, a novel dual-stream hybrid framework that decouples performance and robustness for reliable collaborative perception. 
\item We introduce CIT and LC for efficient feature-level fusion, and PAC for robust object-level correction, decoupling performance from robustness.
\item We conduct comprehensive experiments on multiple datasets. CoRA consistently outperforms over nine powerful methods, setting a new SOTA with minimal communication overhead. 

\end{itemize}

\section{Related Work}

\subsection{Collaborative Perception}
Collaborative perception enables multiple agents to build a comprehensive environmental understanding by exchanging sensory information. This field is broadly categorized into three paradigms based on the fusion stage: early, intermediate, and late fusion. Early fusion is considered impractical due to its prohibitive communication bandwidth requirements \cite{gao2024survey3}. 
Meanwhile, although late fusion is communication-efficient, it suffers from limited performance because fusing detections at the object level discards low-level context\cite{lv2024systematicsuvery5}.
This has led to the prevalence of intermediate fusion, which offers a favorable trade-off between perception accuracy and communication overhead\cite{xu2023v2v4real}. Numerous methods have been proposed within this paradigm. V2X-ViT \cite{xu2022v2xvit} designs specialized attention mechanisms for interaction among heterogeneous agents; CoAlign \cite{coalign} leverages graph neural networks to achieve pose-robust collaboration; DSRC \cite{zhang2024dsrc} employs teacher-student distillation to enhance performance in adverse weather conditions. Despite these advances, purely intermediate fusion exposes a key vulnerability: direct feature-level fusion is susceptible to cross-agent interference. Motivated by this limitation, our work proposes a more robust and practical solution that integrates the high performance of intermediate fusion with the rectifiability of late fusion.

\subsection{Multi-Agent Communication}

To make intermediate fusion practical, numerous strategies have been proposed to reduce communication overhead. One major strategy involves transmitting only the most critical features. 
For instance, sender-centric frameworks like Where2comm \cite{hu2022where2comm} and How2comm \cite{yang2023how2comm} require each agent to select key features to minimize its transmitted volume independently. The second strategy focuses on the compression of holistic features. Methods like CORE \cite{wang2023core} use deep learning to compress the entire feature map, while CodeFilling \cite{hu2024communication} transmits learned codes corresponding to the features for efficiency. Although these methods are effective at reducing the payload from individual agents, they share a common limitation: the communication and computation load on the receiver still scales linearly with the number of collaborators. To address this scalability issue, our work proposes a novel on-demand transmission strategy. This ensures the communication burden remains nearly constant, regardless of the number of participating agents.

\begin{figure*}[t]
    \centering
    \includegraphics[width=0.95\textwidth]{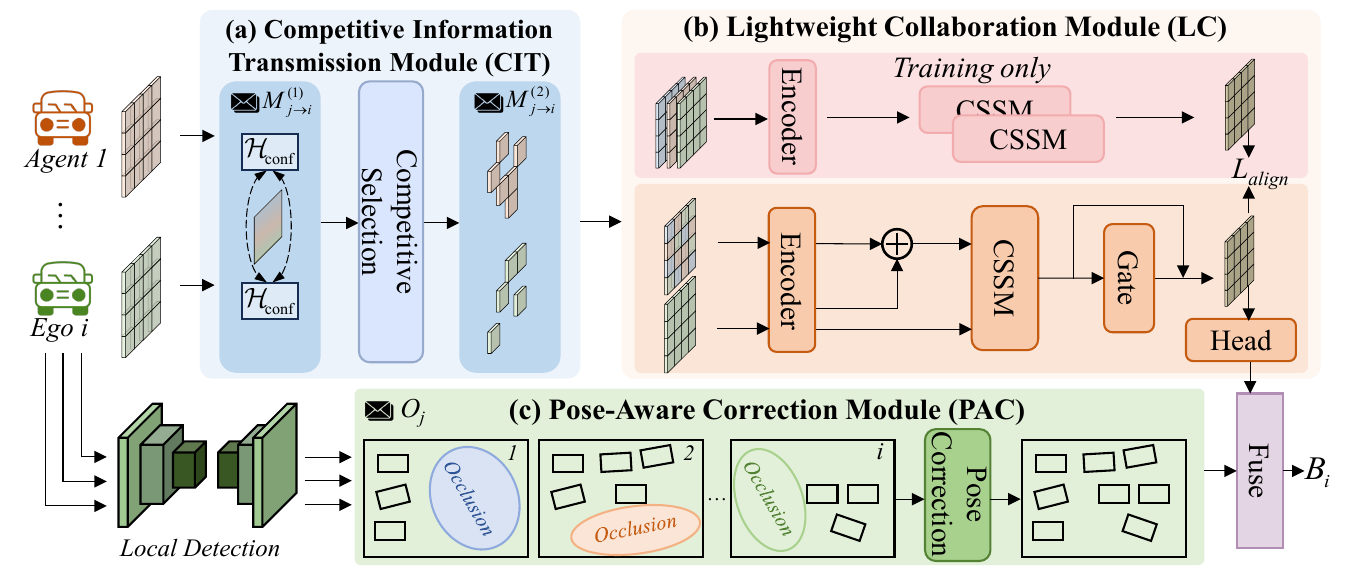} 
    \caption{Overall architecture of the proposed CoRA. CoRA comprises two main branches: a feature-level fusion branch (blue \& orange) and an object-level correction branch (green). The feature branch leverages CIT for competitive feature transmission and LC for post-fusion interaction. The LC's training is guided by dense fused features from other vehicles. Meanwhile, the object-level correction branch transmits collaborator detection results to the ego and corrects them via the PAC module.}
    \label{fig:overall}
\end{figure*}

\section{Methodology}

\subsection{Problem Formulation}

We consider a cooperative scenario involving a set of $N$ agents, including ego and collaborators, denoted by $\mathcal{A} = \{1, 2, ..., N\}$. Each agent $j \in \mathcal{A}$ captures local sensor data $X_j$ and its own pose data $T_j$. To facilitate collaboration under strict bandwidth limits, a collaborator $j$ generates a compact message $M_{j}$ from its local data $X_j$ and transmits this message to the ego $i$. The goal of ego $i$ is to learn a collaborative perception model $\mathcal{F}_{\text{coop}}$. This model takes the ego's own local data $X_i$ and the set of messages $\{M_{j \to i}\}_{j \ne i}$ received from other agents as input, to output the final detection predictions $B_i$. Formally, the problem is to design $\mathcal{F}_{\text{coop}}$:
\begin{align}
B_i = \mathcal{F}_{\text{coop}}(X_i, \{M_{j \to i}\}_{j \in \mathcal{A}}) \quad \text{s.t.} \ |M_{j \to i}| \le C
\end{align}
where the message $M_{j \to i}$ is generated from the data $X_j$ of agent $j$. $|M_{j \to i}|$ represents the communication cost of the message, which is constrained by a budget $C$. However, a fundamental challenge arises in real-world applications: the poses are imperfect. We only have access to noisy poses $\hat{T}_i = T_i \oplus \Delta T_i$, where $\Delta T_i$ is an unknown pose error.

\subsection{Overall Architecture}
As illustrated in Fig. \ref{fig:overall}, our framework for each ego $i$ utilizes a parallel dual-stream architecture to decouple performance from robustness. The feature-level branch first employs a two-stage transmission strategy: collaborators $j$ ($\neq i$) first use Competitive Information Transmission (CIT) module to exchange lightweight confidence maps ($M_{j \to i}^{(1)}$) to identify perception demands, and transmit only sparse features ($M_{j \to i}^{(2)}$). The Lightweight Collaboration (LC) module then fuses these sparse features. LC is designed for information integration with dense feature distillation during training. Concurrently, the object-level branch serves as a compensation mechanism. We propose a novel Pose-Aware Correction (PAC) module to rectify misalignments involving the received detection outputs $O_j$. Finally, the outputs from both branches are fused to produce the final prediction $B_i$.

\subsection{Competitive Information Transmission Module}

Prevailing methods for collaborative perception often adopt a sender-centric communication strategy. For instance, in ERMVP \cite{zhang2024ermvp}, each collaborator compresses and transmits features based on its self-generated or an individual ego's confidence map. While this approach reduces the payload from each agent, the feature volume processed by the ego scales significantly with the number of collaborators. To overcome this bottleneck, we introduce the CIT module, which shifts the paradigm from sender-centric broadcasting to a receiver-centric, on-demand protocol, ensuring a consistently low communication overhead.

CIT operates through a two-stage communication mechanism. In the first stage, each collaborator $j$ processes its sensor data with a local encoder $\mathcal{E}$ to produce a feature map ${F}_j$ $\in \mathbb{R}^{C \times H \times W}$. Subsequently, a confidence head $\mathcal{H}_{\text{conf}}$ generates a map $M_{j \to i}^{(1)} \in \mathbb{R}^{1 \times H \times W}$ as the initial message transmitted to the ego $i$.
\begin{equation}
    M_{j \to i}^{(1)} = \mathcal{H}_{\text{conf}}({F}_j),\quad j \neq i
\end{equation}

Upon receiving $\{M_{j \to i}^{(1)}\}$ from all collaborators, ego $i$ determines the optimal information provider for each spatial location. It first computes its own demand coefficient matrix $D_i$ and uses it to generate a relevance score ${S}_j$ for each collaborator $j$ to enhance regions that the ego cannot perceive:
\begin{align}
D_i &= 1 - \sigma(\mathcal{H}_{\text{conf}}({F}_i)) 
\end{align}
\begin{align}
{S}_j &= D_i \odot M_{j \to i}^{(1)},\quad j \neq i
\end{align}
where $\sigma(\cdot)$ is the sigmoid function and $\odot$ denotes the multiplication by elements. A winner-take-all strategy then performs pixel-wise competition on all relevance maps to determine each pixel's most confident collaborator index.
\begin{equation}
{I}_{\text{win}} = \underset{j \neq i}{\text{argmax}} \left( \text{Stack}(\{{S}_j\}) \right)
\end{equation}
where ${I}_{\text{win}} \in \mathbb{N}^{H \times W}$ assigns each pixel to its best-providing agent. Based on ${I}_{\text{win}}$, the ego $i$ generates a set of binary request masks ${Q}_j$, which are both exclusive and highly sparse.
\begin{equation}
    {Q}_j(h, w) = 
    \begin{cases} 
        1 & \text{if } {I}_{\text{win}}(h, w) = j \\
        0 & \text{otherwise}
    \end{cases}
\end{equation}
where $Q_j \in \{0, 1\}^{1 \times H \times W}$. Each unique request mask ${Q}_j$ is then transmitted back to the corresponding collaborator $j$.

In the second stage, each collaborator $j$ uses its request mask ${Q}_j$ to extract and transmit only the critical sparse features $M_{j \to i}^{(2)} = {F}_j \odot {Q}_j$.

Due to the spatially disjointness of the request masks, ego i directly sums the received features, yielding a consolidated collaborative feature ${F}_{\text{coll}}=\sum_{j \neq i} M_{j \to i}^{(2)}$. This on-demand mechanism transforms the communication burden to a near-constant overhead, thus enhancing system scalability.

\subsection{Lightweight Collaboration Module}

\begin{figure}[!t]
    \centering
    \includegraphics[width=0.45\textwidth]{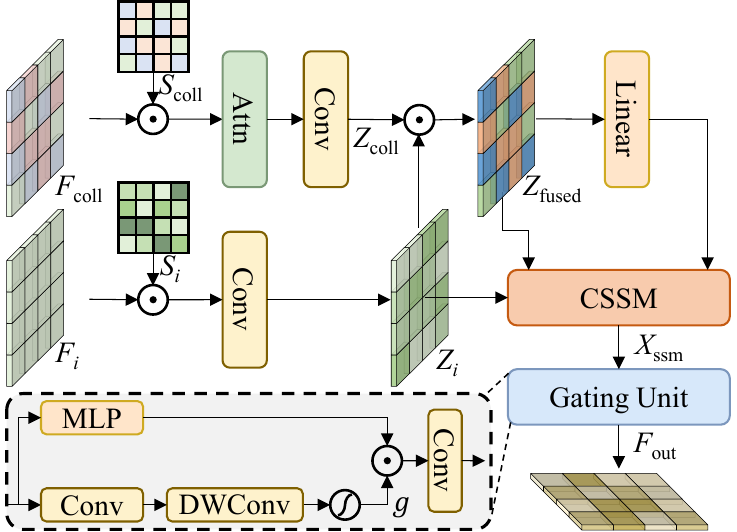}
    \caption{Illustration of the lightweight collaboration module.}
    \label{fig:module2}
\end{figure}

We further introduce the LC module to effectively fuse this collaborative feature ${F}_{\text{coll}}$ with the ego feature ${F}_i$. As depicted in Fig. \ref{fig:module2}, LC leverages Collaborative State Space Model (CSSM) based on Mamba \cite{gu2023mamba} to capture inter-vehicle dependencies and employs a Gating Unit to ensure adaptive feature integration. Meanwhile, a training-only module is used to guide feature distillation.

First, to prioritize features from high-confidence regions, we element-wise multiply ${F}_{\text{coll}}$ and ${F}_i$ with their respective confidence maps, ${S}_{\text{coll}}$ and ${S}_i$. Notably, ${S}_{\text{coll}}$ is generated by aggregating the confidence maps from collaborators.
\begin{align}
    \hat{{F}}_{\text{coll}} &= {F}_{\text{coll}} \odot {S}_{\text{coll}},\quad
    \hat{{F}}_{i} = {F}_{i} \odot {S}_{i}
\end{align}

Next, $\hat{{F}}_{\text{coll}}$ is fed into an attention block \cite{xu2022opv2v} to harmonize the aggregated features from heterogeneous agents, thereby mitigating potential misalignments. Subsequently, both $\hat{{F}}_{\text{coll}}$ and $\hat{{F}}_{i}$ are processed by convolutional operations. The outputs of two branches, ${Z}_{\text{coll}}$ and ${Z}_{i}$, are then fused through element-wise addition, yielding ${Z}_{\text{fused}}$.

CSSM then facilitates efficient inter-vehicle interaction. Similar to modern SSM methods \cite{liu2024vmamba}, CSSM receives three key inputs: ${Z}_{\text{fused}}$ serves as the input matrices, a linear transformation of ${Z}_{\text{fused}}$ parameterizes the step size, and $Z_{i}$ acts as the output matrices. This design enables content-aware contextual modeling, yielding ${X}_{\text{ssm}}$:
\begin{equation}
  {X}_{\text{ssm}} = \text{CSSM}\left({Z}_{\text{fused}},\ \text{Linear}({Z}_{\text{fused}}),\ Z_{i}\right)
\end{equation}

Finally, we employ a Gating Unit for fusion. ${X}_{\text{ssm}}$ is processed to generate a spatial gate ${g} \in \mathbb{R}^{1 \times H \times W}$ via a convolutional operation. Currently, ${X}_{\text{ssm}}$ is fed into MLP, and its output is then modulated by the gate $g$ to get the output ${F}_{\text{out}}$.
\begin{align}
  &{g} = \sigma(\text{DWConv}(\text{Conv}({X}_{\text{ssm}}))) \\
  &{F}_{\text{out}} = \text{Conv}(\text{MLP}({X}_{\text{ssm}}) \odot {g})
\end{align}

\textbf{Feature Distillation.} To compensate for potential information loss from the sparse input ${F}_{\text{coll}}$, we employ a feature distillation mechanism during training. As shown in the top-right corner of the Fig. \ref{fig:overall}, a parallel teacher branch processes complete, non-sparse features to generate a guidance map ${F}_{\text{teacher}}$. ${F}_{\text{out}}$ is encouraged to mimic this target representation via an alignment loss $\mathcal{L}_{\text{align}}$:
\begin{equation}
  \mathcal{L}_{\text{align}} = || {F}_{\text{out}} - {F}_{\text{teacher}} ||_2^2
\end{equation}

\subsection{Pose-Aware Correction Module}

\begin{figure}[!t]
  \centering
  \includegraphics[width=0.47\textwidth]{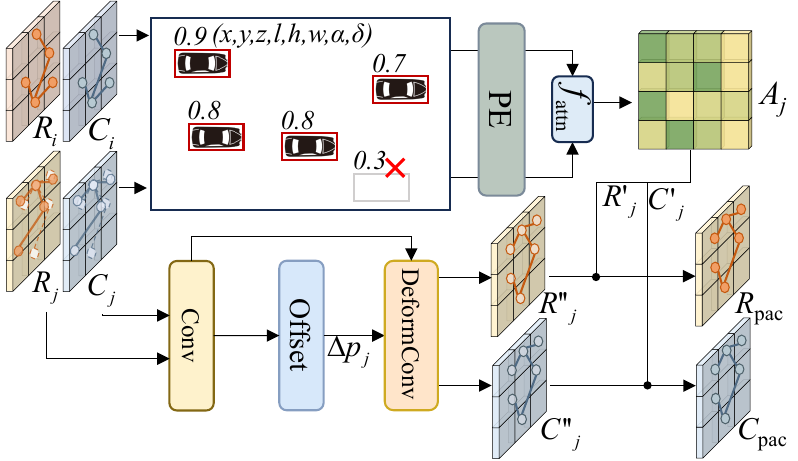}
  \caption{Illustration of the pose-aware correction module.}
  \label{fig:module3}
\end{figure}

To enhance robustness against localization errors inherent in real-world scenarios, we introduce the PAC module, a pose correction module that operates on the object-level outputs from collaborating agents. As shown in Fig. \ref{fig:module3}, PAC receives the detection results from each collaborator $j$, which are represented as classification maps ${C}_j \in \mathbb{R}^{N_\text{cls} \times H \times W}$ and regression maps ${R}_j \in \mathbb{R}^{N_\text{reg} \times H \times W}$. Instead of direct fusion, PAC leverages the ego's results $\{{C}_i, {R}_i\}$ for relevance guidance.

As shown on the top of Fig. \ref{fig:module3}, the alignment process begins by addressing association uncertainty. We first employ convolutional operations to select high-confidence results from collaborator outputs. Next, we generate a descriptor for each detection by incorporating Positional Embedding ($\text{PE}$) of its corresponding box parameters. These descriptors from both ego $i$ and collaborator $j$ are then processed to compute a cross-agent attention map ${A}_j \in \mathbb{R}^{1 \times H \times W}$, which quantifies the matching degree at each spatial location.
\begin{equation}
  {A}_j = \sigma\left(f_{\text{attn}}\left(\text{Concat}(\text{PE}({O}_i), \text{PE}({O}_j))\right)\right)
\end{equation}
where ${O}_i$ and ${O}_j$ denote the detection maps, and $f_{\text{attn}}$ generates the attention map by computing the similarity. The resulting attention map then modulates the collaborator's output, yielding relevance-scored outputs:
\begin{align}
  {C}'_j = {C}_j \odot {A}_j, \quad
  {R}'_j = {R}_j \odot {A}_j
\end{align}

Concurrently, the module performs an explicit semantic misalignment caused by pose errors. We concatenate the detection maps from agents and process them through a shared encoder $f_{\text{offset}}$ to predict a dense 2D offset field $\Delta {p}_j$. This field represents the estimated spatial displacement for each cell of the collaborator's feature map.
\begin{equation}
  \Delta {p}_j = f_{\text{offset}}\left(\text{Concat}({C}_i, {R}_i, {C}_j, {R}_j)\right)
\end{equation}

We then leverage $\Delta {p}_j$ to perform alignment using deformable convolutions \cite{deformconv}. This operation resamples the collaborator's detection maps at the corrected locations specified by $\Delta {p}_j$ to rectify the spatial errors:
\begin{align}
  {C}''_j &= \text{DeformConv}({C}_j, \Delta {p}_j) \\
  {R}''_j &= \text{DeformConv}({R}_j, \Delta {p}_j)
\end{align}

Finally, we fuse these two corrected representations to produce the final collaborator detection maps.

\subsection{Adaptive Final Fusion}

We employ an adaptive fusion mechanism to recalibrate prediction confidence scores of predictions from each branch. We first concatenate classification maps ($C_{\text{lc}}$, $C_{\text{pac}}$) and use a convolutional network to produce uncertainty maps ($U_{\text{lc}}$, $U_{\text{pac}}$), and then adjust initial confidence $\delta$ using associated uncertainty maps $U_{\text{lc/pac}}$ for a recalibrated score.

Finally, predictions from both the feature-level and object-level branches are combined into a single, unified pool. A 3D Non-Maximum Suppression (NMS) operation is then applied to this pool to prune redundant predictions, yielding the ego's definitive collaborative output $B_i$.

\begin{table*}[ht]
\centering
\begin{tabular}{l | c | *{4}{c}}
\hline
Pose error Level $\sigma_t / \sigma_r$ (m/°) & Comm. & 0/0 & 0.2/0.2 & 0.4/0.4 & 0.6/0.6 \\
\hline
Method/Dataset & \multicolumn{5}{c}{OPV2V} \\
\hline
Single & \textbackslash & 0.8078/0.6853 & 0.8078/0.6853 & 0.8078/0.6853 & 0.8078/0.6853 \\
V2VNet \cite{wang2020v2vnet} & 22.74 & 0.9175/0.8221 & 0.9110/0.7566 & 0.8759/0.5779 & 0.7807/0.3950 \\
V2X-ViT \cite{xu2022v2xvit} & 21.35 & 0.9035/0.8119 & 0.8844/0.7336 & 0.8283/0.5711 & 0.7546/0.4266 \\
Where2comm \cite{hu2022where2comm} & 10.74 & 0.8937/0.7889 & 0.8880/0.7159 & 0.8454/0.5439 & 0.7629/0.3904 \\
CoAlign \cite{coalign} & 21.35 & 0.9132/0.8381 & 0.9090/0.7682 & 0.8635/0.5947 & 0.7810/0.4849 \\
ERMVP \cite{zhang2024ermvp} & 6.34 & 0.9139/0.8404 & 0.9085/0.7602 & 0.8588/0.6632 & 0.8303/0.5655 \\
MRCNet \cite{mrcnet} & 21.35 & 0.8775/0.7673 & 0.8509/0.7365 & 0.8063/0.6804 & 0.7330/0.6113 \\
DSRC \cite{zhang2024dsrc} & 21.35 & 0.9183/0.8526 & 0.9133/0.7698 & 0.8663/0.6097 & 0.7753/0.4738 \\
CoSDH \cite{xu2025cosdh} & 6.65 & 0.8952/0.8373 & 0.8589/0.6165 & 0.6922/0.3685 & 0.5453/0.2825 \\
MDD \cite{MDD} & 32.03 & 0.8007/0.6817 & 0.8032/0.6156 & 0.7757/0.4677 & 0.7028/0.3394 \\
\textbf{CoRA} & 3.80 & \textbf{0.9341/0.8658} & \textbf{0.9297/0.7817} & \textbf{0.8858/0.7199} & \textbf{0.8451/0.6544} \\
\hline
Method/Dataset & \multicolumn{5}{c}{DAIR-V2X} \\
\hline
Single & \textbackslash & 0.6250/0.4457 & 0.6250/0.4457 & 0.6250/0.4457 & 0.6250/0.4457 \\
V2VNet \cite{wang2020v2vnet} & 11.71 & 0.6644/0.4037 & 0.6493/0.3879 & 0.6273/0.3675 & 0.5984/0.3508 \\
V2X-ViT \cite{xu2022v2xvit} & 11.42 & 0.7046/0.5240 & 0.6959/0.5197 & 0.6774/0.5119 & 0.6601/0.5036 \\
Where2comm \cite{hu2022where2comm} & 5.71 & 0.6735/0.5317 & 0.6012/0.4493 & 0.6245/0.3301 & 0.4588/0.1650 \\
CoAlign \cite{coalign} & 11.42 & 0.7772/0.6284 & 0.7595/0.5961 & 0.7274/0.5812 & 0.7064/0.5752 \\
ERMVP \cite{zhang2024ermvp} & 3.43 & 0.7042/0.5766 & 0.6740/0.5546 & 0.6544/0.5316 & 0.6228/0.5214 \\
MRCNet \cite{mrcnet} & 11.42 & 0.6648/0.5388 & 0.6506/0.5110 & 0.6148/0.4886 & 0.5935/0.4777 \\
DSRC \cite{zhang2024dsrc} & 11.42 & 0.7852/0.6360 & 0.7698/0.5889 & 0.7085/0.5512 & 0.6754/0.5425 \\
CoSDH \cite{xu2025cosdh} & 3.22 & 0.7675/0.6350 & 0.7472/0.5756 & 0.6925/0.5410 & 0.6607/0.5267 \\
MDD \cite{MDD} & 17.13 & 0.7495/0.5817 & 0.7201/0.5395 & 0.6688/0.5092 & 0.6437/0.4982 \\
\textbf{CoRA} & 2.84 &\textbf{ 0.7858/0.6361} & \textbf{0.7691/0.5977} & \textbf{0.7455/0.5862} & \textbf{0.7312/0.5803} \\
\hline
\end{tabular}
\caption{Comparison of 3D object detection AP@0.5/AP@0.7 under varying magnitudes of pose error on OPV2V and DAIR-V2X. Pose errors are introduced into the pose information, quantified by a translation standard deviation $\sigma_t$ [m] / rotation standard deviation $\sigma_r$ [°]. Comm. represents the average communication reception volume, in MB.}
\label{tab:main_results}
\end{table*}

\section{Experiments}

\subsection{Setup}

\textbf{Dataset.}
To validate the effectiveness of CoRA, we utilize OPV2V \cite{xu2022opv2v} and DAIR-V2X \cite{dairv2x}, two of the most widely used datasets for collaborative perception. Specifically, OPV2V is a large-scale simulated dataset focusing on V2V scenarios, simulating complex multi-vehicle traffic scenarios and communication occlusions. DAIR-V2X is a large-scale real-world dataset that emphasizes V2I scenarios, providing vehicle-side and roadside data in real traffic environments to compensate for single-vehicle perception blind spots. These two datasets cover both simulated and real-world scenarios, as well as the two main collaborative types, enabling a comprehensive and effective evaluation of the model's effectiveness.

\textbf{Experimental Setup.}
We adopted the basic settings from OpenCOOD \cite{xu2022opv2v}. All models were trained on a single RTX 4090 GPU with 24GB. We used the Adam optimizer, set the training epochs to 30, and the batch size to 2. The vehicle communication range was limited to 70 meters. For all models, PointPillar \cite{lang2019pointpillars} was chosen as the backbone network to extract 2D features from point clouds, with each voxel's side length set to 0.4 meters. Performance was evaluated using Average Precision at IoU 0.5 and 0.7 (AP@0.5 / AP@0.7), standard 3D object detection metrics that measure both bounding box accuracy and spatial overlap.

\subsection{Quantitative Evaluation}

\textbf{Comparison of Detection Performance.} As shown in Tab. \ref{tab:main_results}, under ideal conditions, our method achieves state-of-the-art performance on both benchmarks, reaching 86.58\% AP@0.7 on OPV2V and 63.61\% AP@0.7 on DAIR-V2X, respectively. Notably, this top-tier performance is achieved with extremely low communication overheads of only 3.80 MB for OPV2V and 2.84 MB for DAIR-V2X, significantly reducing communication volume by 82.2\% compared to other high-performance methods like DSRC and V2VNet. When low pose error is introduced, our method demonstrates superior robustness, surpassing SOTA performance with 78.17\% AP@0.7 on OPV2V and 59.77\% AP@0.7 on DAIR-V2X. Notably, the incorporation of the post-correction branch might lead to a slight performance perturbation in environments with minimal pose error compared to the intermediate-only branch. This trade-off is acceptable for enhancing efficiency and leveraging the robust advantages of the post-fusion branch.

\begin{figure*}[!t]
  \centering
   \includegraphics[width=1.0\textwidth]{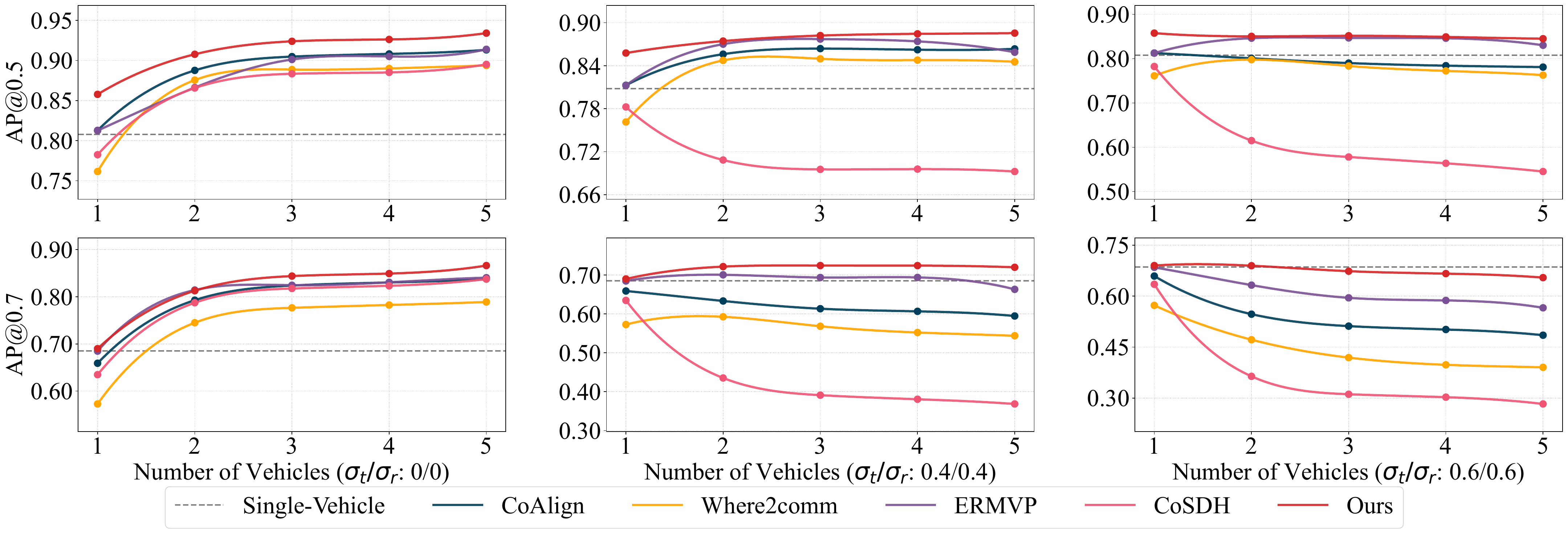} 
    \caption{Comparison across varying numbers of collaborators and pose error levels ($\sigma_t / \sigma_r$).}
    \label{fig:robustness}
\end{figure*}

\textbf{Comparison of Robustness.} As detailed in Tab. \ref{tab:main_results} and Fig. \ref{fig:robustness}, Our model demonstrates exceptional robustness against pose errors. Under the high pose error level of 0.6/0.6 $\sigma_t / \sigma_r$, our approach maintains high accuracy on OPV2V, outperforming strong methods like CoAlign by a significant 17.0\% in AP@0.7. This resilience is further validated in Fig. \ref{fig:robustness}, which shows performance scaling with vehicle count. For instance, while CoSDH exhibits competitive performance under ideal conditions with low communication overhead, its performance degrades gradually as more vehicles with perturbed data are introduced. In sharp contrast, our method consistently maintains its SOTA performance, leveraging collaborators even in noisy multi-agent settings.

\begin{table}[t] 
\setlength{\tabcolsep}{1mm}
\label{tab:addlabel}%
\begin{tabular}{@{} l ccccc @{}} 
\hline
Methods\textbackslash{}Delay & 0ms & 100ms & 200ms & 300ms & 400ms \\
\hline
\multicolumn{6}{@{}l}{AP@0.5} \\ 
\hline
F-cooper & 0.7298 & 0.6288 & 0.4230 & 0.3069 & 0.2746 \\
Where2comm & 0.7629 & 0.6876 & 0.5585 & 0.4650 & 0.4322 \\
CORE & 0.6573 & 0.5443 & 0.3447 & 0.2271 & 0.2077 \\
ERMVP & 0.8303 & 0.6126 & 0.4219 & 0.3007 & 0.2819 \\
\textbf{Ours} & \textbf{0.8433} & \textbf{0.7828} & \textbf{0.6261} & \textbf{0.5088} & \textbf{0.4620} \\
\hline 
\multicolumn{6}{@{}l}{AP@0.7} \\ 
\hline
F-cooper & 0.3189 & 0.2319 & 0.1514 & 0.1462 & 0.1579 \\
Where2comm & 0.3904 & 0.3161 & 0.2333 & 0.2367 & 0.2289 \\
CORE & 0.2289 & 0.1447 & 0.0781 & 0.0709 & 0.0822 \\
ERMVP & 0.5655 & 0.2709 & 0.1705 & 0.1705 & 0.1898 \\
\textbf{Ours} & \textbf{0.6519} & \textbf{0.5683} & \textbf{0.4383} & \textbf{0.3651} & \textbf{0.3519} \\
\hline
\end{tabular}%
\caption{Comparison under varying communication latency with a fixed high pose error ($0.6m/0.6^{\circ}$) on OPV2V.} 
\label{tab:latency_sensitivity}%
\end{table}%

\textbf{Latency Sensitivity Analysis.} Communication latency remains a challenge for communication-efficient methods. We thus evaluate these methods under the challenging combination of latency and pose noise. Tab. \ref{tab:latency_sensitivity} reveals a critical distinction: while the performance of competing methods degrades under delay, CoRA demonstrates far superior resilience. Its ability to maintain AP@0.7 of 0.3651 at 300ms latency highlights its robustness to latency.

\begin{figure}[t]
    \centering
    \includegraphics[width=0.47\textwidth]{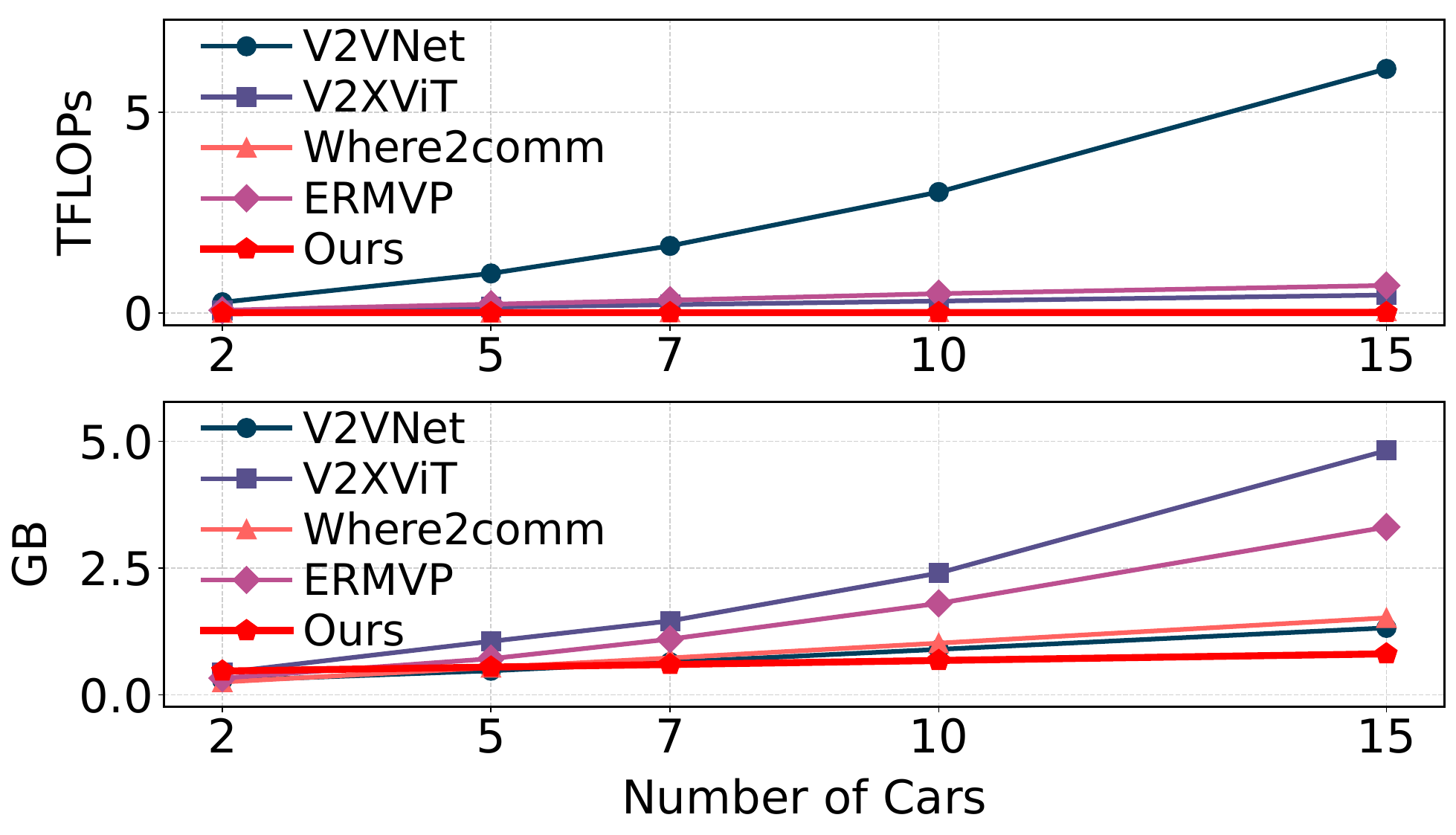} 
    \caption{Comparison of resource consumption across models. Note that we only compare their core modules.}
    \label{fig:cost_analysis}
\end{figure}

\textbf{Computational Cost Analysis.} Fig. \ref{fig:cost_analysis} highlights the efficiency of our core fusion module. CoRA exhibits exceptional scalability. As the number of agents increases, their GFLOPs remain remarkably stable, increasing by a mere 1.4x. This sharply contrasts with the surge of over 22× observed in methods like V2VNet. Our approach offers a significant memory advantage, with usage growing by just 1.72x compared to 4.72x for the next-best method.

\subsection{Ablation Studies}

Tab. \ref{tab:ablation} further illustrates the contributions of the core components CIT, LC, and PAC within our framework. Our baseline is a single-agent object detection model with resnet backbone. Under ideal conditions, the integration of the lightweight feature-level collaboration modules (CIT and LC) significantly boosts performance over the baseline, with gains of 16.06\%/19.02\% in AP@0.5/AP@0.7. In the presence of pose errors, PAC offered enhanced robustness at object-level, yielding an increase of 12.05\%/14.05\% in AP@0.5/AP@0.7 while mitigating the performance degradation typically seen in intermediate fusion. 

\textbf{Effectiveness of CIT.} 
Tab. \ref{tab:idael_compare} compares the performance of a winner-take-all CIT strategy against MaxOut \cite{fcooper} and an alternative Top-2 selection CIT method. The results demonstrate the effectiveness of our strategy.

\begin{table}[t]
\setlength{\tabcolsep}{1mm}
\centering
\begin{tabular}{cclc|cc} 
\hline
$CIT$& $LC$&  $_{+{\mathcal{L}_{\text{align}}}}$&$PAC$& 0/0 & 0.4/0.4 \\ 
\hline
& &  && 0.6250/0.4457 & 0.6250/0.4457 \\ 
\checkmark & &  && 0.7749/0.6036 & 0.6858/0.5184 \\
 \checkmark & \checkmark & & & 0.7826/0.6288&0.6933/0.5287\\ 
\checkmark & \checkmark &  \checkmark && 0.7856/0.6359 & 0.7004/0.5371 \\ 
\checkmark & &  &\checkmark & 0.7786/0.6073 & 0.7270/0.5733 \\ 
\checkmark & \checkmark &  \checkmark &\checkmark & \textbf{0.7858/0.6361} & \textbf{0.7455/0.5862} \\ 
\hline
\end{tabular}%
\caption{Ablation study on DAIR-V2X under ideal/noisy conditions. $\mathcal{L}_{\text{align}}$ denotes the alignment loss introduced within the LC module.} 
\label{tab:ablation}
\end{table}

\begin{table}[t]
  \centering
    \begin{tabular}{c|cc}
    \hline
    Module / Dataset & OPV2V & DAIR-V2X \\
    \hline
    MaxOut & 0.9120/0.8406 & 0.7395/0.5610 \\
   CIT (Top-1)& \textbf{0.9341/0.8661} & \textbf{0.7856/0.6359} \\
   CIT (Top-2)& 0.9325/0.8652 & 0.7839/0.6211 \\
    \hline
    \end{tabular}
  \caption{Performance Comparison of Core Modules.}
  \label{tab:idael_compare}%
\end{table}%

\begin{figure}[!t]
    \centering
    \includegraphics[width=0.45\textwidth]{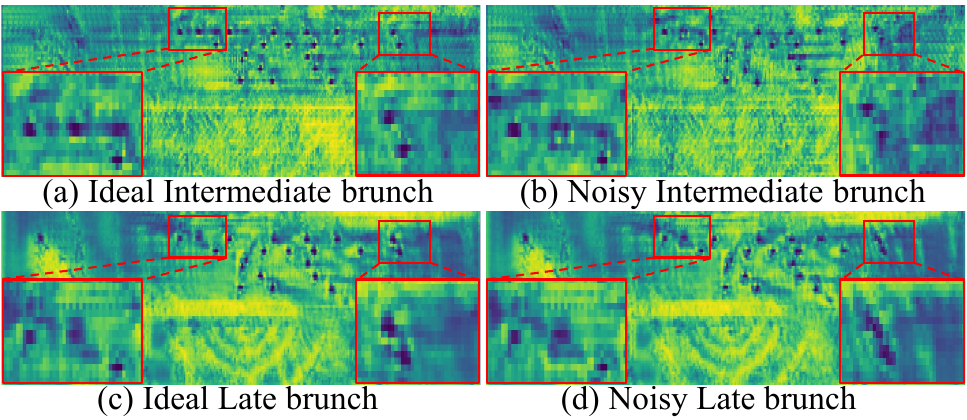} 
    \caption{Visualization of both feature and object-level branches of CoRA under ideal and noisy conditions.}
    \label{fig:vis}
\end{figure}

\subsection{Qualitative Evaluation}

\begin{figure}[!t]
    \centering
    \includegraphics[width=0.47\textwidth]{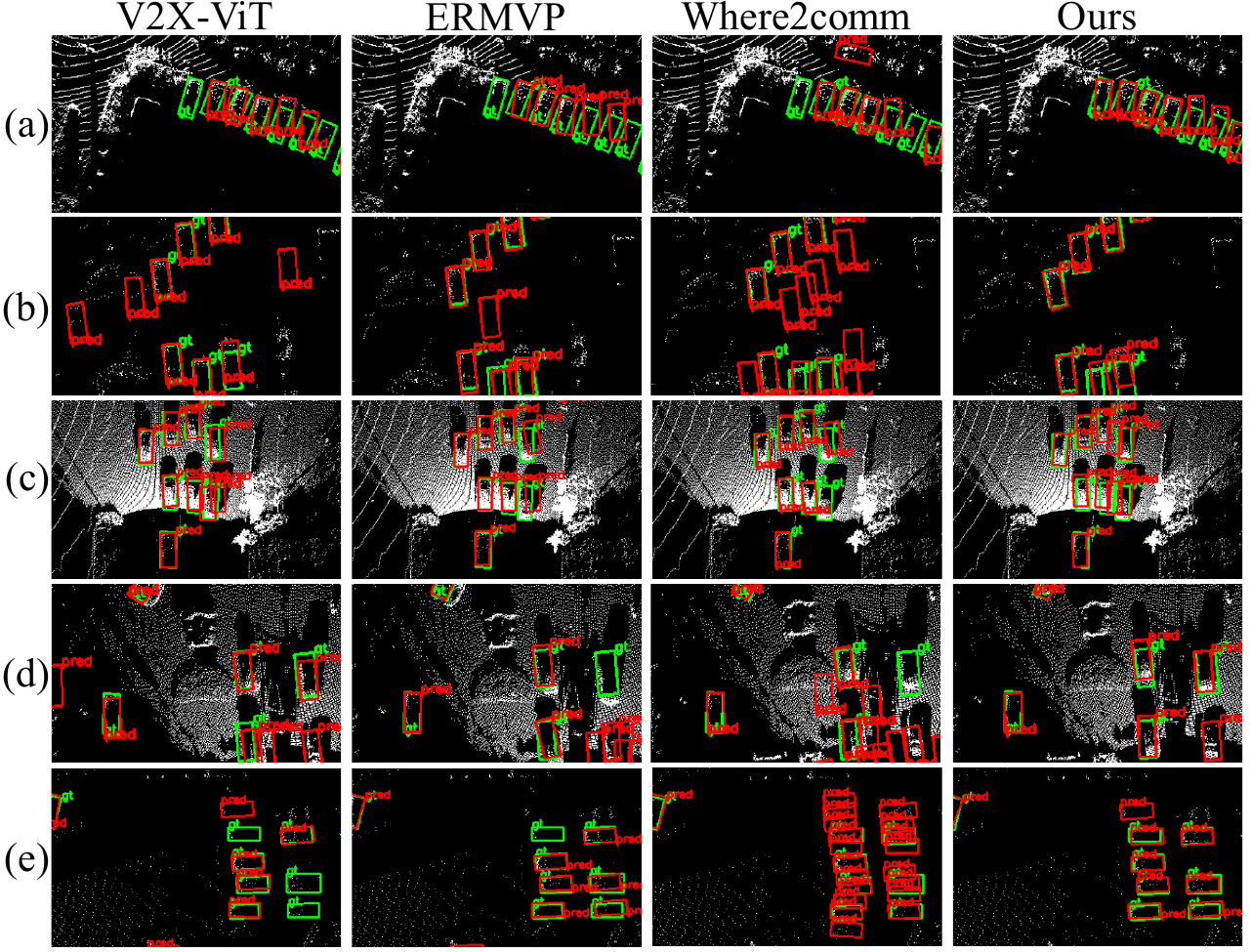} 
    \caption{Visualization of detections in diverse scenarios.}
    \label{fig:vis_small}
\end{figure}

Fig. \ref{fig:vis} shows the complementary design of our method. Under ideal conditions (a), (c), the feature-level branch yields higher-precision localization. In contrast, in the presence of pose error (b), (d), the features of the feature-level branch suffer from severe artifacts, whereas the object-level branch demonstrates high robustness, producing a much cleaner and more coherent representation. Fig. \ref{fig:vis_small} provides visualizations of CoRA's superior detection results in diverse scenarios.

\section{Conclusion}

In this paper, we propose a low-communication and robust collaborative perception method for challenging communication environments. Our approach is distinguished by a novel dual-branch architecture that synergizes intermediate and late fusion. The feature-level branch incorporates a receiver-centric transmission module that ensures near-constant communication overhead, paired with an interaction module for efficient fusion. The object-level branch introduces a novel robust module that corrects misalignments in collaborator detections. Experiments demonstrate that CoRA outperforms current approaches across various scenarios with low communication overhead.

\section{Acknowledgements}

This research is supported in part by the S\&T Program of Hebei Province (Beijing-Tianjin-Hebei Collaborative Innovation Special Program) under Grant 25240701D.

\bibliography{aaai2026}

@InProceedings{	  coalign,
  title		= {Robust collaborative 3D object detection in presence of
		  pose errors},
  author	= {Lu, Yifan and Li, Quanhao and Liu, Baoan and Dianati,
		  Mehrdad and Feng, Chen and Chen, Siheng and Wang, Yanfeng},
  booktitle	= {2023 IEEE International Conference on Robotics and
		  Automation (ICRA)},
  pages		= {4812--4818},
  year		= {2023},
  organization	= {IEEE}
}

@InProceedings{	  dairv2x,
  title		= {DAIR-V2X: A large-scale dataset for vehicle-infrastructure
		  cooperative 3D object detection},
  author	= {Yu, Haibao and Luo, Yizhen and Shu, Mao and Huo, Yiyi and
		  Yang, Zebang and Shi, Yifeng and Guo, Zhenglong and Li,
		  Hanyu and Hu, Xing and Yuan, Jirui and others},
  booktitle	= {2022 IEEE/CVF Conference on Computer Vision and Pattern
		  Recognition (CVPR)},
  pages		= {21361--21370},
  year		= {2022}
}

@InProceedings{	  deformconv,
  title		= {Deformable convolutional networks},
  author	= {Dai, Jifeng and Qi, Haozhi and Xiong, Yuwen and Li, Yi and
		  Zhang, Guodong and Hu, Han and Wei, Yichen},
  booktitle	= {Proceedings of the IEEE International Conference on
		  Computer Vision (ICCV)},
  pages		= {764--773},
  year		= {2017}
}

@InProceedings{	  fcooper,
  title		= {F-cooper: Feature based cooperative perception for
		  autonomous vehicle edge computing system using 3D point
		  clouds},
  author	= {Chen, Qi and Ma, Xu and Tang, Sihai and Guo, Jingda and
		  Yang, Qing and Fu, Song},
  booktitle	= {Proceedings of the 4th ACM/IEEE Symposium on Edge
		  Computing},
  pages		= {88--100},
  year		= {2019}
}

@Article{	  gao2024survey3,
  title		= {A survey of collaborative perception in intelligent
		  vehicles at intersections},
  author	= {Gao, Xin and Zhang, Xinyu and Lu, Yiguo and Huang, Yuning
		  and Yang, Lei and Xiong, Yijin and Liu, Peng},
  journal	= {IEEE Transactions on Intelligent Vehicles (TIV)},
  pages		= {1--20},
  year		= {2024}
}

@InProceedings{	  gu2023mamba,
  title		= {{Mamba: Linear-Time Sequence Modeling with Selective State
		  Spaces}},
  author	= {Gu, Albert and Dao, Tri},
  booktitle	= {First Conference on Language Modeling (CoLM)},
  year		= {2024}
}

@Article{	  hu2022where2comm,
  title		= {Where2comm: Communication-efficient collaborative
		  perception via spatial confidence maps},
  author	= {Hu, Yue and Fang, Shaoheng and Lei, Zixing and Zhong, Yiqi
		  and Chen, Siheng},
  journal	= {Advances in Neural Information Processing Systems
		  (NeurIPS)},
  volume	= {35},
  pages		= {4874--4886},
  year		= {2022}
}

@InProceedings{	  hu2024communication,
  title		= {Communication-efficient collaborative perception via
		  information filling with codebook},
  author	= {Hu, Yue and Peng, Juntong and Liu, Sifei and Ge, Junhao
		  and Liu, Si and Chen, Siheng},
  booktitle	= {2024 IEEE/CVF Conference on Computer Vision and Pattern
		  Recognition (CVPR)},
  pages		= {15481--15490},
  year		= {2024}
}

@InProceedings{	  kekaoxing,
  title		= {A reliability analysis of self-driving vehicles:
		  Evaluating the safety and performance of autonomous driving
		  systems},
  author	= {Pradeep, Aneesh and Bakoev, Mironshokh and Akhroljonova,
		  Nazokat},
  booktitle	= {2023 15th International Conference on Electronics,
		  Computers and Artificial Intelligence (ECAI)},
  pages		= {1--5},
  year		= {2023},
  organization	= {IEEE}
}

@InProceedings{	  lang2019pointpillars,
  title		= {Pointpillars: Fast encoders for object detection from
		  point clouds},
  author	= {Lang, Alex H and Vora, Sourabh and Caesar, Holger and
		  Zhou, Lubing and Yang, Jiong and Beijbom, Oscar},
  booktitle	= {2019 IEEE/CVF Conference on Computer Vision and Pattern
		  Recognition (CVPR)},
  pages		= {12697--12705},
  year		= {2019}
}

@Article{	  late,
  author	= {Arnold, Eduardo and Dianati, Mehrdad and de Temple, Robert
		  and Fallah, Saber},
  journal	= {IEEE Transactions on Intelligent Transportation Systems
		  (TITS)},
  title		= {Cooperative perception for 3D object detection in driving
		  scenarios using infrastructure sensors},
  year		= {2022},
  volume	= {23},
  number	= {3},
  pages		= {1852-1864}
}

@Article{	  liu2024vmamba,
  title		= {Vmamba: Visual state space model},
  author	= {Liu, Yue and Tian, Yunjie and Zhao, Yuzhong and Yu,
		  Hongtian and Xie, Lingxi and Wang, Yaowei and Ye, Qixiang
		  and Jiao, Jianbin and Liu, Yunfan},
  journal	= {Advances in Neural Information Processing Systems
		  (NeurIPS)},
  volume	= {37},
  pages		= {103031--103063},
  year		= {2024}
}

@Article{	  lv2024systematicsuvery5,
  title		= {A systematic literature review of vehicle-to-everything in
		  communication, computation and service scenarios},
  author	= {Lv, Shengnan and Qin, Yong and Gan, Weidong and Xu, Zeshui
		  and Shi, Lefeng},
  journal	= {International Journal of General Systems},
  volume	= {53},
  number	= {7-8},
  pages		= {1042--1072},
  year		= {2024},
  publisher	= {Taylor & Francis}
}

@InProceedings{	  mdd,
  author	= {Huang, Xun and Wang, Jinlong and Xia, Qiming and Chen,
		  Siheng and Yang, Bisheng and Li, Xin and Wang, Cheng and
		  Wen, Chenglu},
  title		= {V2X-R: Cooperative LiDAR-4D radar fusion with denoising
		  diffusion for 3D object detection},
  booktitle	= {Proceedings of the Computer Vision and Pattern Recognition
		  Conference (CVPR)},
  month		= {June},
  year		= {2025},
  pages		= {27390-27400}
}

@InProceedings{	  mrcnet,
  author	= {Hong, Shixin and Liu, Yu and Li, Zhi and Li, Shaohui and
		  He, You},
  booktitle	= {2024 IEEE/CVF Conference on Computer Vision and Pattern
		  Recognition (CVPR)},
  title		= {Multi-Agent Collaborative Perception via Motion-Aware
		  Robust Communication Network},
  year		= {2024},
  volume	= {},
  number	= {},
  pages		= {15301-15310},
  doi		= {10.1109/CVPR52733.2024.01449}
}

@Article{	  siddiqui20215ginterference,
  title		= {Interference management in 5G and beyond networks: A
		  comprehensive survey},
  author	= {Trabelsi, Nessrine and Fourati, Lamia Chaari and Chen,
		  Chung Shue},
  journal	= {Computer Networks},
  volume	= {239},
  pages		= {110159},
  year		= {2024},
  publisher	= {Elsevier}
}

@InProceedings{	  su2023uncertainty,
  title		= {Uncertainty quantification of collaborative detection for
		  self-driving},
  author	= {Su, Sanbao and Li, Yiming and He, Sihong and Han, Songyang
		  and Feng, Chen and Ding, Caiwen and Miao, Fei},
  booktitle	= {2023 IEEE International Conference on Robotics and
		  Automation (ICRA)},
  pages		= {5588--5594},
  year		= {2023},
  organization	= {IEEE}
}

@Article{	  survey1,
  title		= {Explanations in autonomous driving: A survey},
  author	= {Omeiza, Daniel and Webb, Helena and Jirotka, Marina and
		  Kunze, Lars},
  journal	= {IEEE Transactions on Intelligent Transportation Systems
		  (TITS)},
  volume	= {23},
  number	= {8},
  pages		= {10142--10162},
  year		= {2022}
}

@InProceedings{	  wang2020v2vnet,
  title		= {V2VNet: Vehicle-to-vehicle communication for joint
		  perception and prediction},
  author	= {Wang, Tsun-Hsuan and Manivasagam, Sivabalan and Liang,
		  Ming and Yang, Bin and Zeng, Wenyuan and Urtasun, Raquel},
  booktitle	= {Proceedings of the European Conference on Computer Vision
		  (ECCV)},
  pages		= {605--621},
  year		= {2020},
  organization	= {Springer}
}

@InProceedings{	  wang2023core,
  title		= {CORE: Cooperative reconstruction for multi-agent
		  perception},
  author	= {Wang, Binglu and Zhang, Lei and Wang, Zhaozhong and Zhao,
		  Yongqiang and Zhou, Tianfei},
  booktitle	= {Proceedings of the IEEE/CVF International Conference on
		  Computer Vision (ICCV)},
  pages		= {8710--8720},
  year		= {2023}
}

@InProceedings{	  xu2022opv2v,
  title		= {OPV2V: An open benchmark dataset and fusion pipeline for
		  perception with vehicle-to-vehicle communication},
  author	= {Xu, Runsheng and Xiang, Hao and Xia, Xin and Han, Xu and
		  Li, Jinlong and Ma, Jiaqi},
  booktitle	= {2022 International Conference on Robotics and Automation
		  (ICRA)},
  pages		= {2583--2589},
  year		= {2022},
  organization	= {IEEE}
}

@InProceedings{	  xu2022v2xvit,
  title		= {V2X-ViT: Vehicle-to-everything cooperative perception with
		  vision transformer},
  author	= {Xu, Runsheng and Xiang, Hao and Tu, Zhengzhong and Xia,
		  Xin and Yang, Ming-Hsuan and Ma, Jiaqi},
  booktitle	= {Proceedings of the European Conference on Computer Vision
		  (ECCV)},
  year		= {2022}
}

@InProceedings{	  xu2023v2v4real,
  title		= {V2V4Real: A real-world large-scale dataset for
		  vehicle-to-vehicle cooperative perception},
  author	= {Xu, Runsheng and Xia, Xin and Li, Jinlong and Li, Hanzhao
		  and Zhang, Shuo and Tu, Zhengzhong and Meng, Zonglin and
		  Xiang, Hao and Dong, Xiaoyu and Song, Rui and others},
  booktitle	= {2023 IEEE/CVF Conference on Computer Vision and Pattern
		  Recognition (CVPR)},
  pages		= {13712--13722},
  year		= {2023}
}

@InProceedings{	  xu2025cosdh,
  title		= {CoSDH: Communication-efficient collaborative perception
		  via supply-demand awareness and intermediate-late
		  hybridization},
  author	= {Xu, Junhao and Zhang, Yanan and Cai, Zhi and Huang, Di},
  booktitle	= {Proceedings of the Computer Vision and Pattern Recognition
		  Conference (CVPR)},
  pages		= {6834--6843},
  year		= {2025}
}

@Article{	  yang2023how2comm,
  title		= {How2Comm: Communication-efficient and
		  collaboration-pragmatic multi-agent perception},
  author	= {Yang, Dingkang and Yang, Kun and Wang, Yuzheng and Liu,
		  Jing and Xu, Zhi and Yin, Rongbin and Zhai, Peng and Zhang,
		  Lihua},
  journal	= {Advances in Neural Information Processing Systems
		  (NeurIPS)},
  volume	= {36},
  pages		= {25151--25164},
  year		= {2023}
}

@InProceedings{	  yazgan2024survey2,
  title		= {A survey on intermediate fusion methods for collaborative
		  perception categorized by real world challenges},
  author	= {Yazgan, Melih and Graf, Thomas and Liu, Min and Fleck,
		  Tobias and Z{\"o}llner, J Marius},
  booktitle	= {2024 IEEE Intelligent Vehicles Symposium (IV)},
  pages		= {2226--2233},
  year		= {2024},
  organization	= {IEEE}
}

@InProceedings{	  zhang2024dsrc,
  title		= {DSRC: Learning density-insensitive and semantic-aware
		  collaborative representation against corruptions},
  author	= {Zhang, Jingyu and Wang, Yilei and Qian, Lang and Sun, Peng
		  and Li, Zengwen and Jiang, Sudong and Liu, Maolin and Song,
		  Liang},
  booktitle	= {Proceedings of the AAAI Conference on Artificial
		  Intelligence},
  year		= {2025},
  pages		= {9942--9950}
}

@InProceedings{	  zhang2024ermvp,
  title		= {ERMVP: Communication-efficient and collaboration-robust
		  multi-vehicle perception in challenging environments},
  author	= {Zhang, Jingyu and Yang, Kun and Wang, Yilei and Wang,
		  Hanqi and Sun, Peng and Song, Liang},
  booktitle	= {2024 IEEE/CVF Conference on Computer Vision and Pattern
		  Recognition (CVPR)},
  pages		= {12575--12584},
  year		= {2024}
}

@inproceedings{tang2025rocooper,
  title={RoCooper: Robust Cooperative Perception Under Vehicle-to-Vehicle Communication Impairments},
  author={Tang, Tao and Zhang, Chaokun and Chen, Gong and others},
  booktitle={IEEE INFOCOM 2025-IEEE Conference on Computer Communications},
  pages={1--10},
  year={2025},
  organization={IEEE}
}

\end{document}